\pdfoutput=1

\documentclass[sigconf,natbib=true,nonacm]{acmart}
\AtBeginDocument{%
  }

\acmConference{Washington Cunha et al.}{March, 2025}{Brazil}

\usepackage{multirow}
\usepackage{enumitem}
\usepackage{adjustbox}
\usepackage{subcaption}
\usepackage{balance}
\usepackage{relsize}

\begin{document}


\title{
{A thorough benchmark of automatic text classification\\From traditional approaches to large language models}}

\author{Washington Cunha}
\affiliation{%
  \institution{Federal University of Minas Gerais}
  \country{Brazil}
}
\email{washingtoncunha@dcc.ufmg.br}

\author{Leonardo Rocha}
\affiliation{%
  \institution{Federal University of São João del Rei}
  \country{Brazil}
}
\email{lcrocha@ufsj.edu.br}

\author{Marcos André Gonçalves}
\affiliation{%
  \institution{Federal University of Minas Gerais}
  \country{Brazil}
}
\email{mgoncalv@dcc.ufmg.br}

\renewcommand{\shortauthors}{Washington Cunha et al.}


\begin{abstract}

Automatic text classification (ATC)  has experienced remarkable advancements in the past decade, best exemplified by recent small and large language models (SLMs and LLMs),   leveraged by Transformer architectures. Despite recent effectiveness improvements, a comprehensive cost-benefit analysis investigating whether the effectiveness gains of these recent approaches compensate their much higher costs when compared to more traditional text classification approaches such as SVMs and Logistic Regression is still missing in the literature. In this context, this work's main contributions are twofold: (i) we provide a scientifically sound comparative analysis of the cost-benefit of
twelve traditional and recent ATC solutions including five open LLMs, and (ii)  a large benchmark comprising {22 datasets}, including sentiment analysis and topic classification,  with their (train-validation-test) partitions based on folded cross-validation procedures, along with documentation, and code. The release of code, data, and documentation enables the community to replicate experiments and advance the field in a more scientifically sound manner. Our comparative experimental results indicate that LLMs outperform traditional approaches (up to 26\%-7.1\% on average) and SLMs (up to 4.9\%-1.9\% on average) in terms of effectiveness. However, LLMs incur significantly higher computational costs due to fine-tuning, being, on average 590x and 8.5x slower than traditional methods and SLMs, respectively. Results suggests the following recommendations: (1) LLMs for applications that require the best possible effectiveness and can afford the costs; (2) traditional methods such as Logistic Regression and SVM for resource-limited applications or those that cannot afford the cost of tuning large LLMs;  and (3) SLMs like Roberta for near-optimal effectiveness-efficiency trade-off.
Documentation, code, and datasets can be accessed on GitHub -- 
\url{https://github.com/waashk/atcBench/}.
\looseness=-1 

\end{abstract}




\maketitle



\vspace{-0.3cm}
\section{Introduction}

The Web has empowered end users to take an active role in generating content. Consequently, there has been a significant increase in the amount of available data, making its organization a notable challenge. 
Automatic text classification is a well-established task in \textit{Natural Language Processing} and  \textit{Information Retrieval} that offers valuable techniques to address this information overload problem and has become increasingly relevant in many scenarios, such as 
{news categorization, product recommendation, hate speech detection, product features review~\cite{zhang2023prompt,xu2024intelligent,han2021simplest,cunha18}, among others.}\looseness=-1 

Traditional ATC approaches have relied on a variety of supervised machine learning algorithms. Techniques such as Support Vector Machines and Random Forests have long been prominent in this domain. These models leverage carefully hand-crafted features extracted from text (such as TF-IDF~\cite{luhn1957statistical}) to perform classification tasks. Despite their simplicity and efficiency, these traditional models typically require extensive feature engineering and lack the capacity to capture deep semantic relationships within text, especially when compared to deep learning approaches.

Small Language Models (SLMs) such as RoBERTa~\cite{liu2019roberta} and BART~\cite{lewis2019bart}
have obtained outstanding effectiveness across a variety of tasks~\cite{adhocretrieval,documentRanking,alturayeif2023automated,questionAnswering20}.   Large Language Models (LLMs) build on top of these advances by exploiting immense amounts of training data. Many studies indicate that these new models represent the current SOTA in a variety of NLP tasks~\cite{holistic}. However, while the literature emphasizes the benefits of LLMs for specific tasks, such as summarization~\cite{zhang2024benchmarking}, their effectiveness gains for other tasks, such as ATC, the focus of this paper, have sparked debates~\cite{zhao2023survey}, especially due to the lack of rigorous and sound scientific procedures~\cite{cunha21,dacrema19}.

Moreover, achieving those benefits usually requires  vast amounts of training datasets and complex architectures with millions of parameters~\cite{wang2020minilm,10.1145/3634912}.
{Although SLMs and LLMs show some effectiveness in zero-shot scenarios, fine-tuning for specific tasks is essential for optimal performance~\cite{uppaal2023fine}. Thus, there is a clear cost-effectiveness trade-off for these novel approaches. }

Accordingly, the first research question (RQ) we
explore in this paper is: \textbf{RQ1}: \textit{Are LLMs more effective than traditional approaches and SLMs for the ATC task?} To investigate this question, we 
evaluate the performance of 
twelve
traditional approaches, SLMs, and LLMs on 22 datasets, analyzing their effectiveness. 
\looseness=-1

Considering the higher computational costs associated with fine-tuning LLMs to achieve small to moderate improvements over SLMs methods, the second RQ we seek to address is: \textbf{RQ2}: \textit{What is the computational cost of applying LLMs to ATC compared to traditional approaches and SLMs?} We perform a detailed cost analysis by measuring and comparing the computational resources required for 
fine-tuning the methods as well as their estimated carbon emissions.
\looseness=-1

Our experimental results show that LLMs are consistently superior compared to traditional approaches (7\% on average - up to 26\%)
and provide moderate improvements over SLMs 
(up to 4.9\%).
However, these moderate improvements come with 
notably higher computational costs associated with the fine-tuning.  Indeed, LLMs are, respectively,  590x and 8.5x slower than traditional classifiers and SLMs applied to ATC. Based on the application needs, our results suggest: (i) LLMs for the best possible effectiveness, if computational costs can be afforded; (ii) traditional methods for reasonable performance at the lowest cost; and (iii)  SLMs for a near-optimal effectiveness-cost balance.
\looseness=-1

Summarizing, the main contributions of this paper are: (1) an up-to-date, scientifically rigorous 
cost-benefit comparison of traditional approaches, SLMs, and LLMs applied to ATC. Most works in the literature are focused only on effectiveness;
(2) a comprehensive benchmark comprising codes, documentation, results,  datasets, and partitions, enabling direct comparison and further advancements by the community. 
\looseness=-1

\section{Benchmark Scenario and Experiments}

This section presents the benchmark scenario, including datasets, ATC methods, evaluation metrics, and experimental protocol. To ensure reproducibility and future comparison, we provide the documented code for all compared methods, along with datasets, fold divisions, and all obtained results. All these artifacts can be accessed on GitHub:   \url{https://github.com/waashk/atcBench/}.
Compared to prior work~\cite{zhao2024advancing,siino2024text}, our study's novelty lies on the number of conducted experiments, the comparison of traditional, SLMs, and LLMs for ATC, the variety of considered factors (datasets, effectiveness, and cost), and the scientific rigor of our study. .\looseness=-1


\subsection{Datasets}

To evaluate the classifiers, we considered \textbf{22} real-world datasets (Table~\ref{tbl:datasets_stats}) from several sources in two main ATC tasks: i) \textit{topic classification} and ii) \textit{sentiment analysis}. These tasks are recognized as core text classification problems~\cite{sebastiani02,li2022survey,camacho2017role,cunha21}. We also included 3 large datasets to illustrate the classifiers' scalability in big data scenarios. The datasets cover a range of domains, diversity in size, dimensionality, class numbers, document density, and class distribution. 
\looseness=-1

\begin{table}[h]
    \centering
    \resizebox{0.47\textwidth}{!}{
    \begin{tabular}{llrrrrr}
        Task & Dataset       & Size      & Dim. & \# Classes  & Density  & Skewness    \\
        \midrule
\parbox[t]{1mm}{\multirow{11}{*}{\rotatebox[origin=c]{90}{Topic}}}  & DBLP & 38,128 & 28,131 & 10 & 141 & Imbalanced \\
 & Books & 33,594 & 46,382 & 8 & 269 & Imbalanced \\
 & ACM & 24,897 & 48,867 & 11 &  65 & Imbalanced   \\
 & 20NG & 18,846 & 97,401 & 20 &  96 & Balanced    \\
 & OHSUMED & 18,302 & 31,951 & 23 & 154 & Imbalanced   \\
 & Reuters90 & 13,327 & 27,302 & 90 & 171 & Extremely   Imbalanced \\
 & WOS-11967 & 11,967 & 25,567 & 33 & 195 & Balanced \\
 & WebKB & 8,199 & 23,047 & 7 & 209 & Imbalanced   \\
& Twitter & 6,997 & 8,135 & 6 & 28 & Imbalanced \\ 
 & TREC & 5,952 & 3,032 & 6 & 10 & Imbalanced \\
 & WOS-5736 & 5,736 & 18,031 & 11 & 201 & Balanced \\ \hline
\parbox[t]{1mm}{\multirow{8}{*}{\rotatebox[origin=c]{90}{Sentiment}}}  & SST1 & 11,855 & 9,015 & 5 & 19 & Balanced         \\
 & pang\_movie & 10,662 & 17,290 & 2 & 21 & Balanced        \\
 & Movie Review & 10,662 & 9,070 & 2 & 21 & Balanced   \\
 & vader\_movie & 10,568 & 16,827 & 2 & 19 & Balanced        \\
 & MPQA & 10,606 & 2,643 & 2  & 3 & Imbalanced         \\
 & Subj & 10,000 & 10,151 & 2 & 24 & Balanced         \\
 & SST2 & 9,613 & 7,866 & 2 & 19 & Balanced         \\
 & yelp\_reviews & 5,000 & 23,631 & 2 & 132 & Balanced        \\
\hline
\parbox[t]{1mm}{\multirow{3}{*}{\rotatebox[origin=c]{90}{Large}}} & AGNews & 127,600 & 39,837 & 4 & 37 & Balanced \\
& Yelp\_2013 & 335,018 & 62,964 & 6 & 152 & Imbalanced \\
& MEDLINE &	 860,424 & 125,981 & 7 & 77 & Extremely Imbalanced \\

        \bottomrule
    \end{tabular}
    }
    \caption{Datasets Statistics}
    \label{tbl:datasets_stats}
\end{table}

\subsection{ATC Methods}

In~\cite{cunha21}, the authors evaluate the performance of various traditional machine learning models for ATC. In our analysis, we 
decided to focus on methods that consistently demonstrated strong performance across different datasets and scenarios. 
Therefore, we considered the following top-3 traditional models: \textbf{Support Vector Machines (SVM)}, \textbf{Logistic Regression (LR)}, and \textbf{Random Forests (RF)}.\looseness=-1

{In~\cite{Cunha2023A}, the authors compare the effectiveness of seven SLMs in the context of ATC. For our analysis, we focused on the methods that achieved the highest absolute Macro-F1 score at least once. 
Consequently, we considered the following models:} \textbf{RoBERTa}~\cite{liu2019roberta}, \textbf{BERT}~\cite{bert}, \textbf{BART}~\cite{lewis2019bart}, and \textbf{XLNet}~\cite{yang2019xlnet}, notably SOTA in ATC.
\looseness=-1

As previously stated, our objective is to study and compare several ATC methods to create a comprehensive benchmark of the current up-to-date SOTA. To accomplish this, we focus on open-source LLMs, as closed-source and proprietary LLMs (e.g., ChatGPT) are black boxes that prevent us from understanding how they were trained or their internal structure. 
We chose {four} SOTA open-source~\cite{spirling2023open} models:
\textbf{DeepSeek}~\cite{guo2025deepseek}, \textbf{LLaMA}~\cite{touvron2023}, \textbf{Mistral}~\cite{jiang2023mistral}, and \textbf{BloomZ}~\cite{muennighoff2022crosslingual}. 
In their fine-tuned versions, these LLMs have consistently demonstrated robust performance across a range of domains~\cite{de2024strategy}.
Next, we provide a brief description of these models:\looseness=-1

\noindent - \textbf{SVM}~\cite{cortes1995support} 
identifies the near-optimal hyperplane to separate classes by maximizing the margin from critical samples.

\noindent - \textbf{RF}~\cite{breiman2001random} is an ensemble method that constructs multiple decision trees using bootstrap samples and random feature splits. 

\noindent - \textbf{LR}~\cite{cabrera1994logistic} is a linear model that predicts the probability of an instance belonging to a class using the logistic function, adjusting feature weights to minimize the error between predictions and actual labels.\looseness=-1

\noindent - \textbf{BERT}~\cite{bert} employs a bidirectional transformer encoder. BERT predicts missing words using \textit{masked language model} for 15\% of the words and \textit{Next sentence prediction} to assess sentence continuity.

\noindent - \textbf{RoBERTa}~\cite{liu2019roberta} improved BERT by enhancing tuning with: 
more data, batch sizes up to 32K, longer training time, larger sequence (up to 512), pre-training objective and dynamic mask generation.

\noindent - \textbf{BART}~\cite{lewis2019bart} is a denoising autoencoding model that merges a bidirectional encoder with an autoregressive decoder. 
\looseness=-1

\noindent - \textbf{XLNet}~\cite{yang2019xlnet}: Unlike BERT's independence assumption for predicted tokens, XLNet utilizes a permutation language modeling objective, blending autoregressive and autoencoder advantages.

\noindent - \textbf{LLaMA}~\cite{touvron2023} is a family of autoregressive LLMs released by Meta AI, optimized for multilingual dialogue applications.
In this work, we focus on two LLaMA versions: 2 (7B) and 3.1 (8B).

\noindent - \textbf{Mistral}-7B-v0.3~\cite{jiang2023mistral} 
uses grouped-query attention for faster inference and sliding window attention for 
reduced costs.

\noindent - \textbf{BloomZ}~\cite{muennighoff2022crosslingual} extended the fine-tuning for BLOOM and mT5 multilingual language models applied to a cross-lingual mixture, 
overcoming several models in unseen tasks and languages.
\looseness=-1

\noindent - \textbf{DeepSeek}-R1-Distill-Llama-8B~\cite{guo2025deepseek} trains using cold-start data and reinforcement learning without initial supervised fine-tuning.\looseness=-1


\vspace{-0.25cm}
\subsection{Experimental Setup}

We evaluated the classifiers on the basis of effectiveness and runtime. 
For the topic and sentiment datasets, we used 10-fold cross-validation, while for larger datasets, we opted for 5-fold due to procedural costs. Effectiveness was assessed using Macro-F1\cite{Sokolova} due to dataset skewness. We performed a paired t-test at a 95\% confidence level to evaluate statistical significance with Bonferroni correction~\cite{bonferroni} to account for multiple tests. {To analyze cost-effectiveness, we also evaluated the total time to build each model, which comprises the average (by fold) time to train the model and apply it to the test set.} All experiments were conducted on the same machine: an Intel Core i7-5820K (12-Threads, 3.30GHz), 64GB RAM, and a NVIDIA RTX 3090 (24GB).\looseness=-1

\begin{table*}[!h]
\centering
       	\centering
    \includegraphics[width=0.93\textwidth]{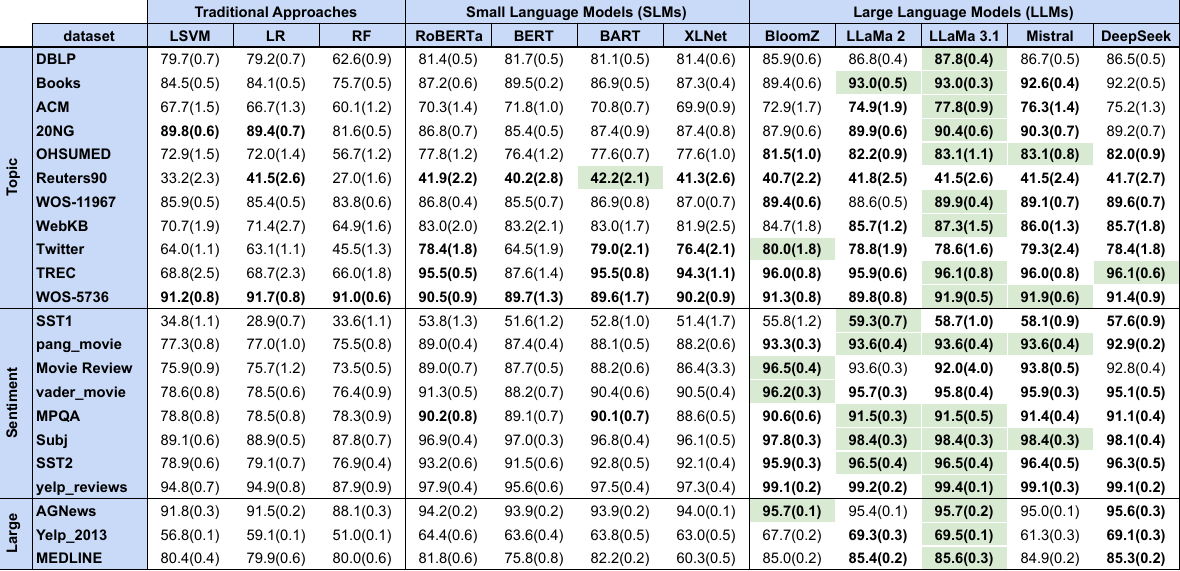}
 \caption{Effectiveness: Macro-F1 and Confidence Interval (CI) of each classifier. Bold values mean statistical ties (t-test with Bonferroni correction), and green backgrounds highlight the best (i.e., higher absolute value) performance per dataset.}
    \label{tab:effect}
\end{table*}

We defined the hyperparameters\footnote{All hyperparameters for all models are provided on Github in Hydra~\cite{Yadan2019Hydra} format.} search for the traditional methods based on~\cite{Cunha2020}. For representation, we adopted TF-IDF for its simplicity and efficiency in the ATC domain. Alternatives~\cite{Cunha2020}, such as static embedding, can result in effectiveness losses compared to TF-IDF (+SVM), and contextual embeddings can be 1.5x to 31.1x slower than TF-IDF. For SVM, we varied the C parameter among [0.125, 0.5, 1, 2, 8, 32, 128, 512, 2048, 8192]. For LR, we varied the C among [0.001, 0.01, 0.1, 1]. For RF, we evaluated the number of generated trees among [50, 100, 200] and the max\_feat to [0.08, 0.15, 0.30]. \looseness=-1

Given the large number of hyperparameters to be tuned for the SLMs and LLMs, performing a grid search with cross-validation is not feasible for all of them. Therefore, for 
the SLMs, we applied the methodology in~\cite{Cunha2023B}. We fixed the initial learning rate as 5e-5, the max number of epochs as 20, and 5 epochs as patience. Finally, we performed a grid search on max\_len (128 and 256) and batch\_size (16 and 32) since these specified values directly impact efficiency and effectiveness.
Based on~\cite{de2024strategy}, for the LLMs, we adopted 4-bit quantization, with QLoRA and PEFT for efficient fine-tuning on our on-premise machine,
fixing
the initial learning rate as 2e-4, the number of epochs and the batch\_size as 4, and the max\_len as 256.

\section{Experimental Results}\label{sec:expresults}

\subsection{Effectiveness}\label{sec:effect}

As shown in Table~\ref{tab:effect}, LLMs consistently demonstrated superior effectiveness in most datasets when compared to traditional approaches and SLMs. Specifically, LLMs achieve the highest Macro-F1 score in 21 out of the 22 considered datasets. Even when focusing solely on LLaMa 3.1, the results remain impressive, as it secures the highest scores in 17 (out of 22) datasets and is statistically comparable to the top-performing model in the remaining ones.  Mistral and LLaMA 2 (18 wins in 22 datasets) and DeepSeek (17 wins in  22 datasets) follow this trend.

Traditional models (LSVM, LR, and RF) performed substantially worse than LLMs on most datasets. For example, in Reuters90, LSVM achieved a Macro-F1 score of 33.2, well below  LLaMA, Mistral, and DeepSeek (on average, 41.5).
 
Despite their lower performance on average, traditional models exhibited competitive scores in some cases, e.g., in 20NG and WOS-5736, where simpler representations sufficed. However, in general, they were outperformed by both SLMs and LLMs. In fact, on average, LLMs outperformed traditional methods by approximately 7.2\% (up to 26.04\% -- SST1) in terms of Macro-F1. 

Finally, LLMs also outperformed SLMs such as RoBERTa, BERT, and BART in most cases. For example, in yelp\_reviews, LLaMa 3.1 achieved the best performance (99.4), being statistically better  than all SLMs, which varied between 95.6 (BERT) and 97.9 (RoBERTa). In a few datasets, though, e.g., Reuters, Twitter, TREC, WOS-5736, and MPQA,  LLMs and SLMs were statistically tied, suggesting that SLMs are  competitive on specific tasks. Perhaps even more important it is the magnitude of the gains which were \textbf{not high}. 
On average, LLMs outperform SLMs by only  1.94\% (min=0.31\% and max=4.86\%).
\looseness=-1

Answering  \textbf{RQ1},  LLMs are often superior to traditional approaches and SLMs for ATC tasks, achieving the highest Macro-F1 scores in most cases. Nevertheless, regarding SLMs, the (average) magnitude of the gains are low.  

\begin{table*}[!h]
\centering
        \includegraphics[width=0.97\textwidth]{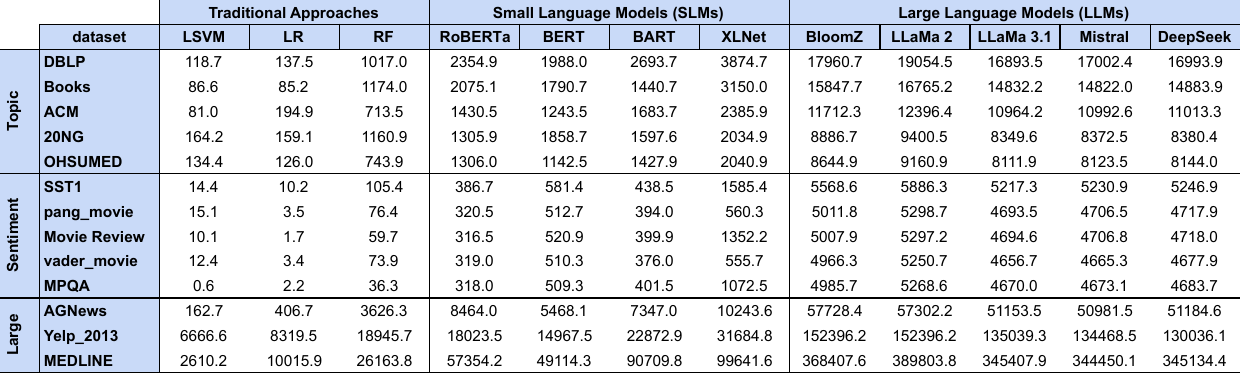}
 \caption{{Total application cost (training + test time in seconds) of the classifiers' application in each dataset. Due to length constraints, we only show a selection of datasets; conclusions for others remain similar. All results are accessible on GitHub.}}
    \label{tab:time}
\end{table*}

\subsection{Efficiency}

In this section, we consider the computational and environmental costs of traditional approaches, SLMs, and LLMs for ATC when evaluating their overall utility.
As shown in Table~\ref{tab:time}, in all datasets, traditional methods exhibit the lowest application time by orders of magnitude compared to both SLMs and LLMs.
For example, in the DBLP dataset, the total cost for traditional methods ranges from around 2 minutes per fold (LSVM) to 17 minutes (RF). Other datasets, such as  SST1, show times as low as one second (MPQA). 
Even in larger datasets such as MEDLINE, the highest computational time for traditional methods (about 1 hour by RF) is still significantly lower than that of SLMs
and LLMs (on average 102 hours).\looseness=-1

 SLMs require significantly more time than traditional methods. For instance, on DBLP, the cost ranges from 
33 minutes (BERT) to more than 
one hour (65 minutes in XLNet). 
SLMs increase 2x-80x the cost compared to traditional methods (14x on average). 
{The discrepancy is even more pronounced with large datasets like Medline. For the sake of magnitude comparison, XLNet alone incurs a cost of over 28 hours per fold, exceeding the total application time of all traditional approaches combined (approximately 24 hours).}
\looseness=-1

Finally, LLMs incur in dramatically higher application costs.
For example, in the DBLP dataset, LLaMa 3.1 requires around 5 hours per fold, significantly surpassing SLMs 
(around 46 minutes) 
and traditional methods (between 2-17 minutes). The disparity is particularly extreme in larger datasets such as AGNews, where LLaMa 3.1, DeepSeek, and other LLMs consume approximately 14 hours each, contrasting to the mere 3 minutes taken by LSVM. In sum, LLMs are, on average, \textbf{590 times computationally costlier} than traditional methods and 
\textbf{8.5} times slower than SLMs.
\looseness=-1

Answering \textbf{RQ2}, while LLMs provide state-of-the-art performance in ATC (as demonstrated in effectiveness results - Section~\ref{sec:effect}), their computational costs are substantially higher compared to SLMs and traditional methods.


\begin{figure}[!ht]
\centering
\includegraphics[width=0.45\textwidth]{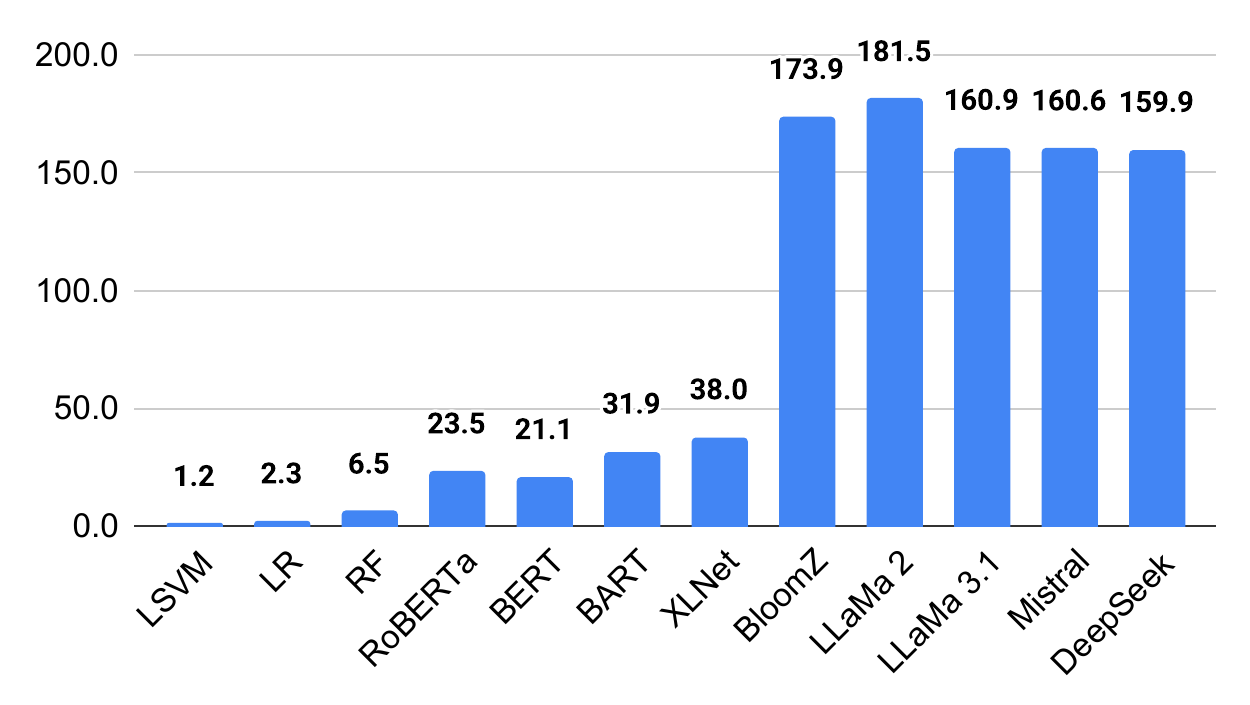}
\caption{CO2e: the equivalent amount of carbon dioxide  (in~kg) generated by the classification models’ fine-tuning. }
    \label{fig:co2e}
\end{figure}

\paragraph{\textbf{Carbon emissions (CO2e):} }
Following the methodology presented in~\cite{lannelongue2021green}, we analyzed the emission of CO2 -- an estimated measure of greenhouse gases -- converted to their equivalent amount of carbon dioxide, which is generated during the models' fine-tuning.
\looseness=-1

Our analysis of CO2e underscores significant disparities in the environmental impact of these approaches. In Figure~\ref{fig:co2e}, we can notice traditional machine learning models
exhibit a much lower carbon footprint (between 1.2 to 6.5kg of CO2e). These models are computationally inexpensive and environmentally friendly, offering a cost-alternative for ATC tasks. In contrast, recent solutions leveraging SLMs and LLMs demonstrate substantially higher carbon emissions. Specifically, SLMs 
generate moderate CO2e emissions, reflecting the high computational costs associated with fine-tuning. Even more pronounced emissions (orders of magnitude higher) are observed for advanced LLMs
with emissions reaching around 
170kg
each.
\looseness=-1

In total, we performed experiments worth about 
6400 
hours of computation for this paper. We estimate that this computation resulted in about 
961 
kg of CO2e emissions. To put this into perspective, these emissions are equivalent to driving a passenger car for 
{5500 km or taking thirteen 
flights from Paris to London.
Thus, by extending our benchmark, the community can significantly reduce COe emissions while effectively comparing novel ATC proposals.
\looseness=-1

Ultimately, these findings emphasize the need for more sustainable proposals in developing novel technologies~\cite{pasin2025quantumclef} as well as wisely choosing among the approaches that have already been proposed~\cite{ipmclaudio} to better accommodate specific needs while seeking effectiveness-efficiency trade-off.\looseness=-1


\begin{figure*}[!ht]
    \centering
    \begin{subfigure}{0.32\textwidth}
        \centering
        \includegraphics[height=1.8in]{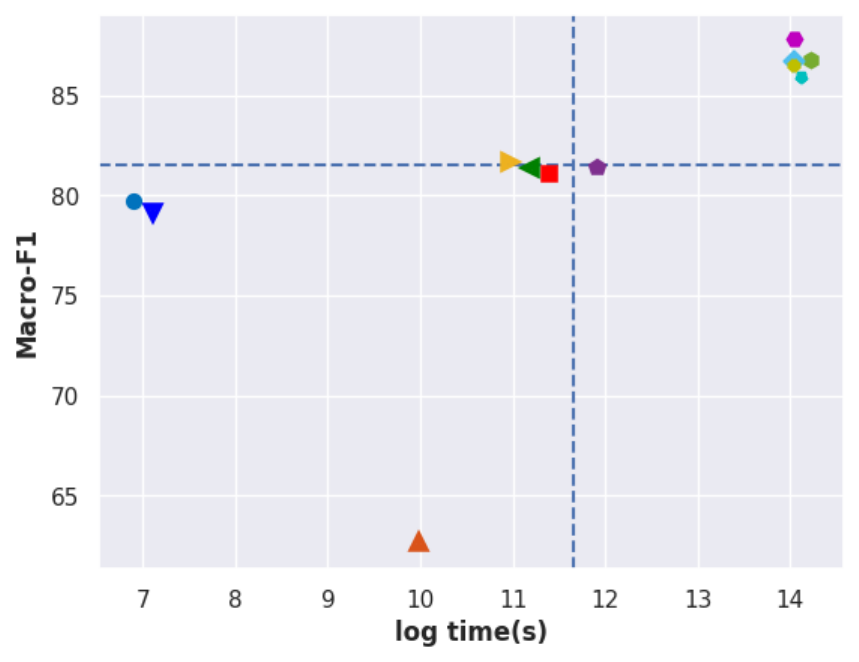}
        \caption{DBLP}
    \end{subfigure}%
    ~
    \begin{subfigure}{0.32\textwidth}
        \centering
        \includegraphics[height=1.8in]{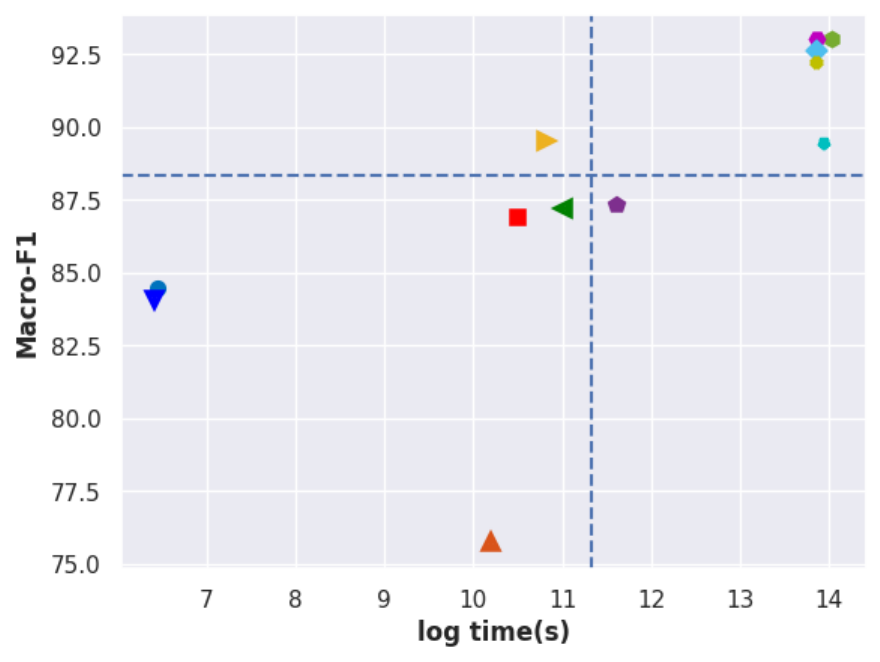}
        \caption{Books}
    \end{subfigure}%
    ~
    \begin{subfigure}{0.32\textwidth}
        \centering
        \includegraphics[height=1.8in]{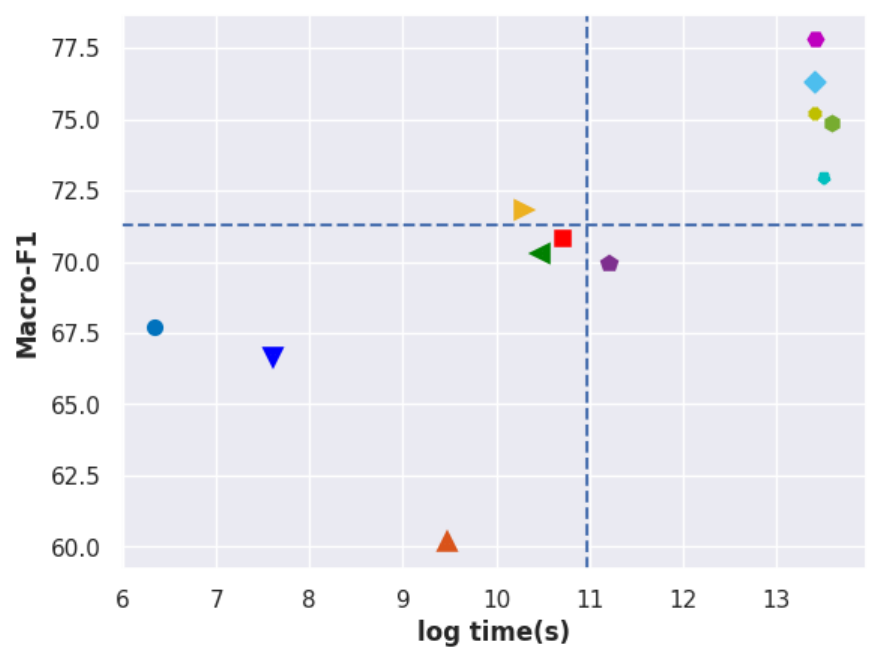}
        \caption{ACM}
    \end{subfigure}%

    \begin{subfigure}{0.32\textwidth}
        \centering
        \includegraphics[height=1.8in]{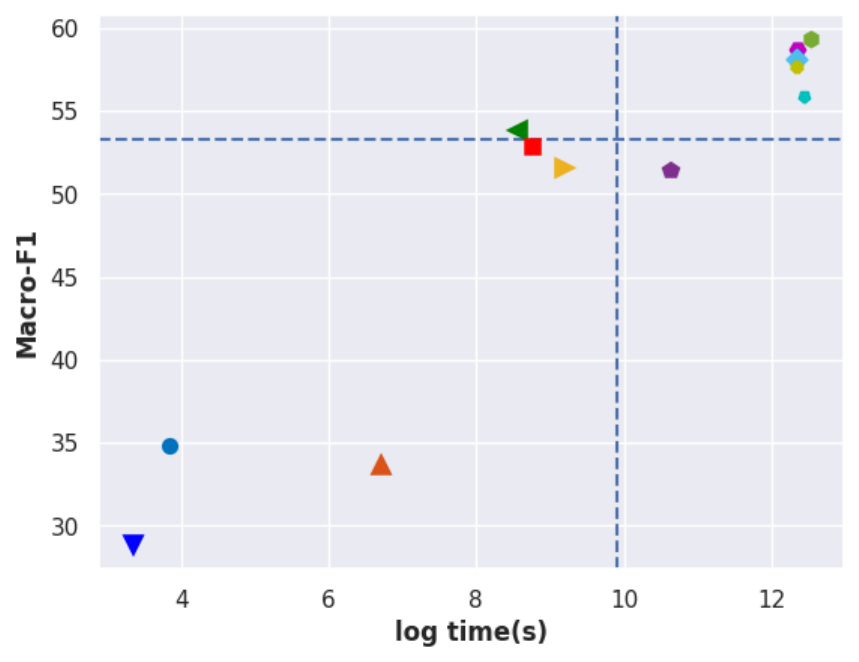}
        \caption{SST1}
    \end{subfigure}%
    ~
    \begin{subfigure}{0.32\textwidth}
        \centering
        \includegraphics[height=1.8in]{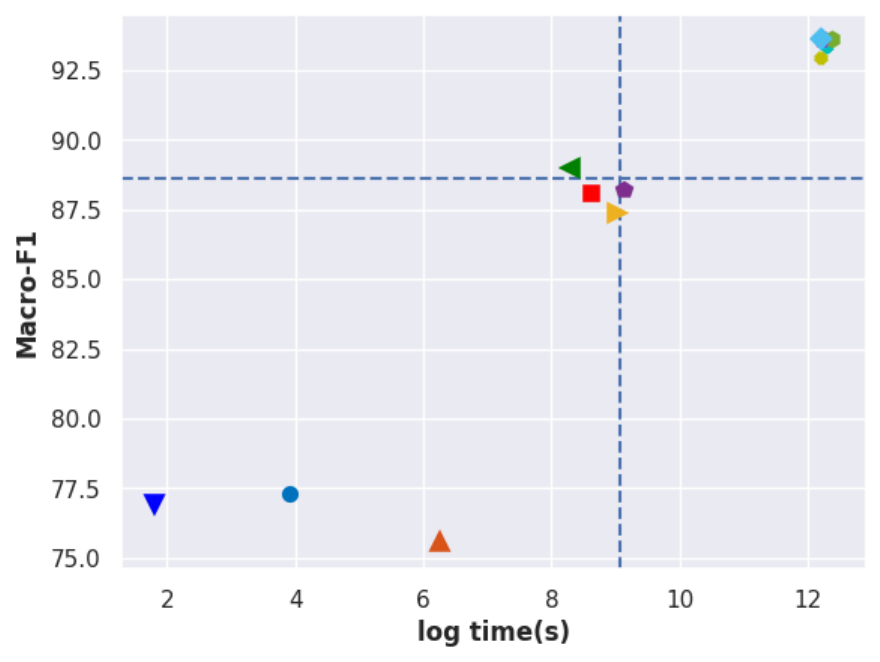}
        \caption{pang\_movie}
    \end{subfigure}%
    ~
    \begin{subfigure}{0.32\textwidth}
        \centering
        \includegraphics[height=1.8in]{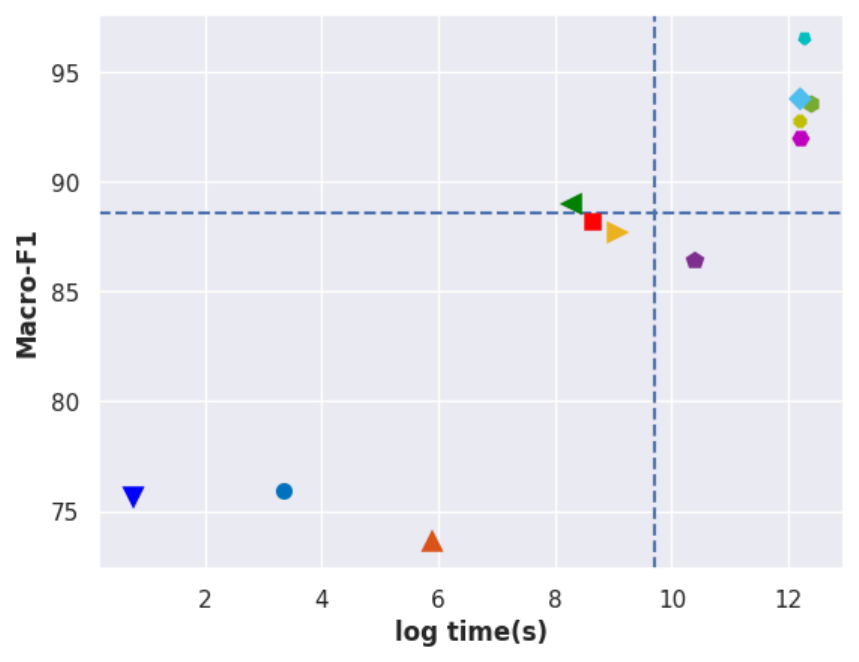}
        \caption{Movie Review}
    \end{subfigure}%

    \begin{subfigure}{0.32\textwidth}
        \centering
        \includegraphics[height=1.8in]{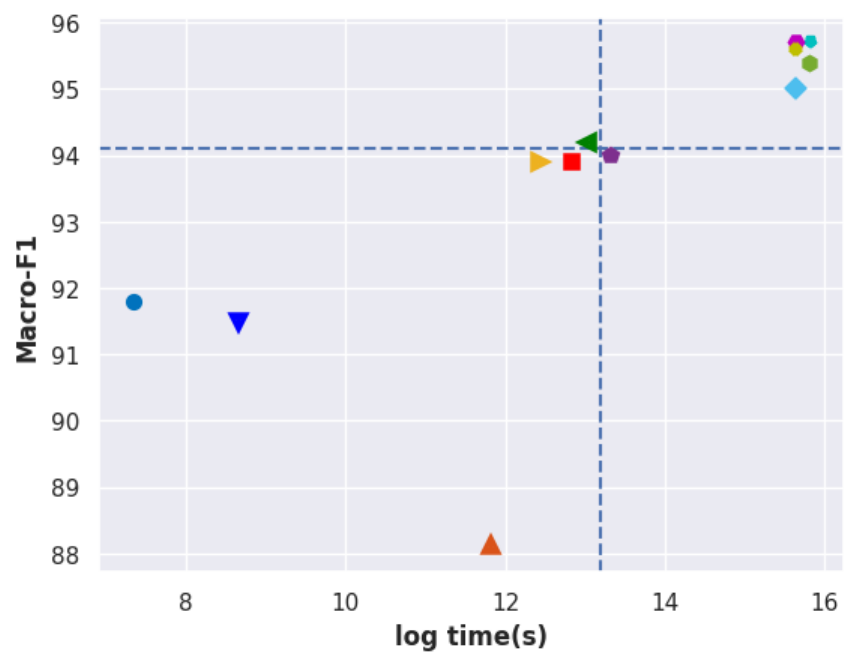}
        \caption{AGNews}
    \end{subfigure}%
    ~
    \begin{subfigure}{0.32\textwidth}
        \centering
        \includegraphics[height=1.8in]{resource/plots/pang_movie.pdf}
        \caption{Yelp\_2013}
    \end{subfigure}%
    ~
    \begin{subfigure}{0.32\textwidth}
        \centering
        \includegraphics[height=1.8in]{resource/plots/mr.pdf}
        \caption{MEDLINE}
    \end{subfigure}%
    
    \begin{subfigure}{0.7\textwidth}
        \centering
        \includegraphics[width=\textwidth]{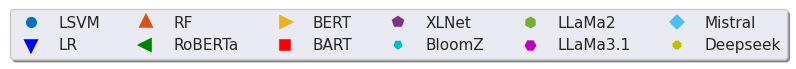}
    \end{subfigure}%

    \caption{Cost (log2 training time)-Effectiveness (MacroF1) Trade-off for each dataset.}
    \label{sum}
\end{figure*}

\subsection{Effectiveness-Efficiency Trade-Off}

We analyze here issues related to the cost-effectiveness tradeoff of the methods at training and prediction. Figure~\ref{sum} presents the Macro-F1 results versus time (for train and prediction) for all evaluated methods across different datasets. The aim is to visualize how effective each method is in relation to the total time spent on training and prediction. On the y-axis, we measure the Macro-F1 score of each method, while the x-axis displays the logarithm of the time (in seconds). We utilize the logarithm of time so that more efficient methods appear on the left side of the axis. The dotted lines indicate the median values for both axes (log(time) and Macro-F1). Methods located in the first quadrant outperform the median in both effectiveness and efficiency. The second quadrant contains methods that are highly effective but also more time-consuming. Methods in the third quadrant are below the median on both time and effectiveness. Lastly, the fourth quadrant includes methods that are the least effective in terms of Macro-F1 but are faster than the median. In sum, the more to the left and the higher, the better the cost-effectiveness tradeoff.\looseness=-1

The results in Figure~\ref{sum} show that top-performer RoBERTa is also moderately fast to train.  It is always in the first quadrant, usually close to the middle. The other best overall performer, BERT presents a very good cost-effectiveness tradeoff -- it is usually in the first quadrant (DBLP, Books, ACM) or very close to it. In absolute terms, the overall slowest methods, consistently across all datasets, are LLaMA, Mistral, and DeepSeek (closer to upper right). While the fastest ones include LSVM, LR, and RF,  which interchange positions in the top three in the different datasets. Comparing these methods, there is a huge advantage for the traditional methods from the performance efficiency perspective. Evaluating the absolute times achieved by the methods, we observe that the runtimes differ by orders of magnitude. However, although much slower, LLMs usually obtain the top effectiveness results.

\paragraph*{Summary}





We summarize the obtained results and discuss their implications, providing a conclusive and actionable analysis for both practitioners and researchers. As demonstrated in our experiments, in terms of effectiveness (RQ1), LLMs are often superior to traditional approaches and SLMs for ATC tasks, achieving the highest Macro-F1 scores in most cases.
However, their computational costs (RQ2) are substantially higher compared to SLMs and traditional methods. Therefore, if an application demands the highest level of effectiveness and there are no budget constraints or environmental concerns, it is advisable to adopt LLMs~\cite{kaddour2023challenges}. On the other hand, if you are looking for a cost-effective model that still offers reasonable prediction accuracy, traditional models are a practical and highly suitable choice~\cite{pasinictir24,pasin2022quantum}. Lastly, if your application requires a near-optimal balance between effectiveness and efficiency, SLMs, such as RoBERTa for ATC, are the best alternative, providing strong effectiveness while significantly incurring fewer computational costs, specially compared to LLMs~\cite{cunha24}.

\section{Conclusion}\label{sec:con}

We conducted a comprehensive cost-benefit analysis of twelve traditional and recent approaches to Automatic Text Classification, implementing an extensive experimental framework.
Our results demonstrate that while LLMs perform better in terms of effectiveness, they incur significantly higher computational costs than traditional models and SLMs. This trade-off highlights the importance of aligning model selection with specific application requirements, whether prioritizing best effectiveness (LLMs), cost efficiency (traditional models), or a near-optimal balance (SLMs). For future work, we intend to extend these results by including new models, such as Falcon~\cite{gao2024falcon}, phi\cite{abdin2024phi}, Gemma~\cite{team2024gemma}, and Qwen~\cite{bai2023qwen}. We also intend to integrate SLMs and LLMs based on document difficulty, seeking the effectiveness-efficiency trade-off. Finally, we encourage the research community to build upon our work by extending this comparative benchmark with other ATC methods as well as additional datasets, metrics, and cost-effectiveness dimensions.\looseness=-1

\paragraph*{\textbf{Acknowledgements}} This work was partially supported by CNPq, CAPES, INCT-TILD-IAR, FAPEMIG, AWS, Google, NVIDIA, CIIA-Saúde, and FAPESP. 

\balance
\bibliographystyle{ACM-Reference-Format}
\bibliography{main}


\begin{thebibliography}{52}


\ifx \showCODEN    \undefined \def \showCODEN     #1{\unskip}     \fi
\ifx \showDOI      \undefined \def \showDOI       #1{#1}\fi
\ifx \showISBNx    \undefined \def \showISBNx     #1{\unskip}     \fi
\ifx \showISBNxiii \undefined \def \showISBNxiii  #1{\unskip}     \fi
\ifx \showISSN     \undefined \def \showISSN      #1{\unskip}     \fi
\ifx \showLCCN     \undefined \def \showLCCN      #1{\unskip}     \fi
\ifx \shownote     \undefined \def \shownote      #1{#1}          \fi
\ifx \showarticletitle \undefined \def \showarticletitle #1{#1}   \fi
\ifx \showURL      \undefined \def \showURL       {\relax}        \fi
\providecommand\bibfield[2]{#2}
\providecommand\bibinfo[2]{#2}
\providecommand\natexlab[1]{#1}
\providecommand\showeprint[2][]{arXiv:#2}

\bibitem[Abdin et~al\mbox{.}(2024)]%
        {abdin2024phi}
\bibfield{author}{\bibinfo{person}{Marah Abdin}, \bibinfo{person}{Jyoti Aneja}, \bibinfo{person}{Harkirat Behl}, \bibinfo{person}{S{\'e}bastien Bubeck}, \bibinfo{person}{Ronen Eldan}, \bibinfo{person}{Suriya Gunasekar}, \bibinfo{person}{Michael Harrison}, \bibinfo{person}{Russell~J Hewett}, \bibinfo{person}{Mojan Javaheripi}, \bibinfo{person}{Piero Kauffmann}, {et~al\mbox{.}}} \bibinfo{year}{2024}\natexlab{}.
\newblock \showarticletitle{Phi-4 technical report}.
\newblock \bibinfo{journal}{\emph{arXiv preprint arXiv:2412.08905}} (\bibinfo{year}{2024}).
\newblock


\bibitem[Alturayeif et~al\mbox{.}(2023)]%
        {alturayeif2023automated}
\bibfield{author}{\bibinfo{person}{Nouf Alturayeif}, \bibinfo{person}{Hamoud Aljamaan}, {and} \bibinfo{person}{Jameleddine Hassine}.} \bibinfo{year}{2023}\natexlab{}.
\newblock \showarticletitle{An automated approach to aspect-based sentiment analysis of apps reviews using machine and deep learning}.
\newblock \bibinfo{journal}{\emph{Automated Software Engineering}} \bibinfo{volume}{30}, \bibinfo{number}{2} (\bibinfo{year}{2023}), \bibinfo{pages}{30}.
\newblock


\bibitem[Bai et~al\mbox{.}(2023)]%
        {bai2023qwen}
\bibfield{author}{\bibinfo{person}{Jinze Bai}, \bibinfo{person}{Shuai Bai}, \bibinfo{person}{Yunfei Chu}, \bibinfo{person}{Zeyu Cui}, \bibinfo{person}{Kai Dang}, \bibinfo{person}{Xiaodong Deng}, \bibinfo{person}{Yang Fan}, \bibinfo{person}{Wenbin Ge}, \bibinfo{person}{Yu Han}, \bibinfo{person}{Fei Huang}, {et~al\mbox{.}}} \bibinfo{year}{2023}\natexlab{}.
\newblock \showarticletitle{Qwen technical report}.
\newblock \bibinfo{journal}{\emph{arXiv preprint arXiv:2309.16609}} (\bibinfo{year}{2023}).
\newblock


\bibitem[Breiman(2001)]%
        {breiman2001random}
\bibfield{author}{\bibinfo{person}{Leo Breiman}.} \bibinfo{year}{2001}\natexlab{}.
\newblock \showarticletitle{Random forests}.
\newblock \bibinfo{journal}{\emph{Machine learning}}  \bibinfo{volume}{45} (\bibinfo{year}{2001}), \bibinfo{pages}{5--32}.
\newblock


\bibitem[Cabrera(1994)]%
        {cabrera1994logistic}
\bibfield{author}{\bibinfo{person}{AF Cabrera}.} \bibinfo{year}{1994}\natexlab{}.
\newblock \showarticletitle{Logistic regression analysis in higher education: An applied perspective}.
\newblock \bibinfo{journal}{\emph{Higher education: Handbook of theory and research}} (\bibinfo{year}{1994}).
\newblock


\bibitem[Camacho-Collados and Pilehvar(2018)]%
        {camacho2017role}
\bibfield{author}{\bibinfo{person}{Jose Camacho-Collados} {and} \bibinfo{person}{Mohammad~Taher Pilehvar}.} \bibinfo{year}{2018}\natexlab{}.
\newblock \showarticletitle{On the Role of Text Preprocessing in Neural Network Architectures: An Evaluation Study on Text Categorization and Sentiment Analysis}. In \bibinfo{booktitle}{\emph{Proceedings of the EMNLP Workshop}}.
\newblock


\bibitem[Cortes and Vapnik(1995)]%
        {cortes1995support}
\bibfield{author}{\bibinfo{person}{Corinna Cortes} {and} \bibinfo{person}{Vladimir Vapnik}.} \bibinfo{year}{1995}\natexlab{}.
\newblock \showarticletitle{Support vector machine}.
\newblock \bibinfo{journal}{\emph{Machine learning}} \bibinfo{volume}{20}, \bibinfo{number}{3} (\bibinfo{year}{1995}), \bibinfo{pages}{273--297}.
\newblock


\bibitem[Cunha et~al\mbox{.}(2020)]%
        {Cunha2020}
\bibfield{author}{\bibinfo{person}{Washington Cunha}, \bibinfo{person}{S{\'e}rgio Canuto}, \bibinfo{person}{Felipe Viegas}, \bibinfo{person}{Thiago Salles}, \bibinfo{person}{Christian Gomes}, \bibinfo{person}{Vitor Mangaravite}, \bibinfo{person}{Elaine Resende}, \bibinfo{person}{Thierson Rosa}, \bibinfo{person}{Marcos~Andr{\'e} Gon{\c{c}}alves}, {and} \bibinfo{person}{Leonardo Rocha}.} \bibinfo{year}{2020}\natexlab{}.
\newblock \showarticletitle{Extended pre-processing pipeline for text classification: On the role of meta-feature representations, sparsification and selective sampling}.
\newblock \bibinfo{journal}{\emph{Information Processing \& Management}} \bibinfo{volume}{57}, \bibinfo{number}{4} (\bibinfo{year}{2020}), \bibinfo{pages}{102263}.
\newblock


\bibitem[Cunha et~al\mbox{.}(2023a)]%
        {Cunha2023B}
\bibfield{author}{\bibinfo{person}{Washington Cunha}, \bibinfo{person}{Celso Fran{\c{c}}a}, \bibinfo{person}{Guilherme Fonseca}, \bibinfo{person}{Leonardo Rocha}, {and} \bibinfo{person}{Marcos~Andr{\'e} Gon{\c{c}}alves}.} \bibinfo{year}{2023}\natexlab{a}.
\newblock \showarticletitle{An Effective, Efficient, and Scalable Confidence-Based Instance Selection Framework for Transformer-Based Text Classification}. In \bibinfo{booktitle}{\emph{Proceedings of the 46th International ACM SIGIR}}. \bibinfo{pages}{665--674}.
\newblock


\bibitem[Cunha et~al\mbox{.}(2021)]%
        {cunha21}
\bibfield{author}{\bibinfo{person}{Washington Cunha}, \bibinfo{person}{Vítor Mangaravite}, \bibinfo{person}{Christian Gomes}, \bibinfo{person}{Sérgio Canuto}, \bibinfo{person}{Elaine Resende}, \bibinfo{person}{Cecilia Nascimento}, \bibinfo{person}{Felipe Viegas}, \bibinfo{person}{Celso França}, \bibinfo{person}{Wellington~Santos Martins}, \bibinfo{person}{Jussara~M. Almeida}, \bibinfo{person}{Thierson Rosa}, \bibinfo{person}{Leonardo Rocha}, {and} \bibinfo{person}{Marcos~André Gonçalves}.} \bibinfo{year}{2021}\natexlab{}.
\newblock \showarticletitle{On the cost-effectiveness of neural and non-neural approaches and representations for text classification: A comprehensive comparative study}.
\newblock \bibinfo{journal}{\emph{Information Processing \& Management}} \bibinfo{volume}{58}, \bibinfo{number}{3} (\bibinfo{year}{2021}), \bibinfo{pages}{102481}.
\newblock
\showISSN{0306-4573}


\bibitem[Cunha et~al\mbox{.}(2024)]%
        {cunha24}
\bibfield{author}{\bibinfo{person}{Washington Cunha}, \bibinfo{person}{Alejandro Moreo}, \bibinfo{person}{Andrea Esuli}, \bibinfo{person}{Fabrizio Sebastiani}, \bibinfo{person}{Leonardo Rocha}, {and} \bibinfo{person}{Marcos Andr{\'e}~Gon{\c{c}}alves}.} \bibinfo{year}{2024}\natexlab{}.
\newblock \showarticletitle{A Noise-Oriented and Redundancy-Aware Instance Selection Framework}.
\newblock \bibinfo{journal}{\emph{ACM TOIS}} (\bibinfo{year}{2024}).
\newblock


\bibitem[Cunha et~al\mbox{.}(2023b)]%
        {Cunha2023A}
\bibfield{author}{\bibinfo{person}{Washington Cunha}, \bibinfo{person}{Felipe Viegas}, \bibinfo{person}{Celso Fran{\c{c}}a}, \bibinfo{person}{Thierson Rosa}, \bibinfo{person}{Leonardo Rocha}, {and} \bibinfo{person}{Marcos~Andr{\'e} Gon{\c{c}}alves}.} \bibinfo{year}{2023}\natexlab{b}.
\newblock \showarticletitle{A Comparative Survey of Instance Selection Methods applied to NonNeural and Transformer-Based Text Classification}.
\newblock \bibinfo{journal}{\emph{Comput. Surveys}} (\bibinfo{year}{2023}).
\newblock


\bibitem[de~Andrade et~al\mbox{.}(2024)]%
        {de2024strategy}
\bibfield{author}{\bibinfo{person}{Claudio de Andrade}, \bibinfo{person}{Washington Cunha}, \bibinfo{person}{Davi Reis}, \bibinfo{person}{Adriana~Silvina Pagano}, \bibinfo{person}{Leonardo Rocha}, {and} \bibinfo{person}{Marcos~Andr{\'e} Gon{\c{c}}alves}.} \bibinfo{year}{2024}\natexlab{}.
\newblock \showarticletitle{A Strategy to Combine 1stGen Transformers and Open LLMs for Automatic Text Classification}.
\newblock \bibinfo{journal}{\emph{arXiv preprint arXiv:2408.09629}} (\bibinfo{year}{2024}).
\newblock


\bibitem[{de Andrade} et~al\mbox{.}(2023)]%
        {ipmclaudio}
\bibfield{author}{\bibinfo{person}{Claudio~M.V. {de Andrade}}, \bibinfo{person}{Fabiano~M. Belém}, \bibinfo{person}{Washington Cunha}, \bibinfo{person}{Celso França}, \bibinfo{person}{Felipe Viegas}, \bibinfo{person}{Leonardo Rocha}, {and} \bibinfo{person}{Marcos~André Gonçalves}.} \bibinfo{year}{2023}\natexlab{}.
\newblock \showarticletitle{On the class separability of contextual embeddings representations – or The classifier does not matter when the (text) representation is so good!}
\newblock \bibinfo{journal}{\emph{IP\&M}} (\bibinfo{year}{2023}).
\newblock


\bibitem[Devlin et~al\mbox{.}(2019)]%
        {bert}
\bibfield{author}{\bibinfo{person}{Jacob Devlin}, \bibinfo{person}{Ming-Wei Chang}, \bibinfo{person}{Kenton Lee}, {and} \bibinfo{person}{Kristina Toutanova}.} \bibinfo{year}{2019}\natexlab{}.
\newblock \showarticletitle{Bert: Pre-training of deep bidirectional transformers for language understanding}. In \bibinfo{booktitle}{\emph{Proceedings of naacL-HLT}}, Vol.~\bibinfo{volume}{1}. Minneapolis, Minnesota, \bibinfo{pages}{2}.
\newblock


\bibitem[Ferrari~Dacrema et~al\mbox{.}(2019)]%
        {dacrema19}
\bibfield{author}{\bibinfo{person}{Maurizio Ferrari~Dacrema}, \bibinfo{person}{Paolo Cremonesi}, {and} \bibinfo{person}{Dietmar Jannach}.} \bibinfo{year}{2019}\natexlab{}.
\newblock \showarticletitle{Are we really making much progress? A worrying analysis of recent neural recommendation approaches}. In \bibinfo{booktitle}{\emph{Proceedings of the 13th ACM conference on recommender systems}}. \bibinfo{pages}{101--109}.
\newblock


\bibitem[Formal et~al\mbox{.}(2024)]%
        {10.1145/3634912}
\bibfield{author}{\bibinfo{person}{Thibault Formal}, \bibinfo{person}{Carlos Lassance}, \bibinfo{person}{Benjamin Piwowarski}, {and} \bibinfo{person}{St\'{e}phane Clinchant}.} \bibinfo{year}{2024}\natexlab{}.
\newblock \showarticletitle{Towards Effective and Efficient Sparse Neural Information Retrieval}.
\newblock \bibinfo{journal}{\emph{ACM Trans. Inf. Syst.}} \bibinfo{volume}{42}, \bibinfo{number}{5}, Article \bibinfo{articleno}{116} (\bibinfo{date}{apr} \bibinfo{year}{2024}).
\newblock
\showISSN{1046-8188}
\urldef\tempurl%
\url{https://doi.org/10.1145/3634912}
\showDOI{\tempurl}


\bibitem[Gao et~al\mbox{.}(2024)]%
        {gao2024falcon}
\bibfield{author}{\bibinfo{person}{Xiangxiang Gao}, \bibinfo{person}{Weisheng Xie}, \bibinfo{person}{Yiwei Xiang}, {and} \bibinfo{person}{Feng Ji}.} \bibinfo{year}{2024}\natexlab{}.
\newblock \showarticletitle{Falcon: Faster and Parallel Inference of Large Language Models through Enhanced Semi-Autoregressive Drafting and Custom-Designed Decoding Tree}.
\newblock \bibinfo{journal}{\emph{arXiv preprint arXiv:2412.12639}} (\bibinfo{year}{2024}).
\newblock


\bibitem[Guo et~al\mbox{.}(2025)]%
        {guo2025deepseek}
\bibfield{author}{\bibinfo{person}{Daya Guo}, \bibinfo{person}{Dejian Yang}, \bibinfo{person}{Haowei Zhang}, \bibinfo{person}{Junxiao Song}, \bibinfo{person}{Ruoyu Zhang}, \bibinfo{person}{Runxin Xu}, \bibinfo{person}{Qihao Zhu}, \bibinfo{person}{Shirong Ma}, \bibinfo{person}{Peiyi Wang}, \bibinfo{person}{Xiao Bi}, {et~al\mbox{.}}} \bibinfo{year}{2025}\natexlab{}.
\newblock \showarticletitle{DeepSeek-R1: Incentivizing Reasoning Capability in LLMs via Reinforcement Learning}.
\newblock \bibinfo{journal}{\emph{arXiv preprint arXiv:2501.12948}} (\bibinfo{year}{2025}).
\newblock


\bibitem[Han et~al\mbox{.}(2021)]%
        {han2021simplest}
\bibfield{author}{\bibinfo{person}{Xiao Han}, \bibinfo{person}{Yuqi Liu}, {and} \bibinfo{person}{Jimmy Lin}.} \bibinfo{year}{2021}\natexlab{}.
\newblock \showarticletitle{The simplest thing that can possibly work:(pseudo-) relevance feedback via text classification}. In \bibinfo{booktitle}{\emph{Proceedings of the 2021 ACM SIGIR International Conference on Theory of Information Retrieval}}.
\newblock


\bibitem[Hochberg(1988)]%
        {bonferroni}
\bibfield{author}{\bibinfo{person}{Yosef Hochberg}.} \bibinfo{year}{1988}\natexlab{}.
\newblock \showarticletitle{A sharper Bonferroni procedure for multiple tests of significance}.
\newblock \bibinfo{journal}{\emph{Biometrika}} \bibinfo{volume}{75}, \bibinfo{number}{4} (\bibinfo{year}{1988}).
\newblock


\bibitem[Jiang et~al\mbox{.}(2023)]%
        {jiang2023mistral}
\bibfield{author}{\bibinfo{person}{Albert~Q Jiang}, \bibinfo{person}{Alexandre Sablayrolles}, \bibinfo{person}{Arthur Mensch}, \bibinfo{person}{Chris Bamford}, \bibinfo{person}{Devendra~Singh Chaplot}, \bibinfo{person}{Diego de~las Casas}, \bibinfo{person}{Florian Bressand}, \bibinfo{person}{Gianna Lengyel}, \bibinfo{person}{Guillaume Lample}, \bibinfo{person}{Lucile Saulnier}, {et~al\mbox{.}}} \bibinfo{year}{2023}\natexlab{}.
\newblock \showarticletitle{Mistral 7B}.
\newblock \bibinfo{journal}{\emph{arXiv preprint arXiv:2310.06825}} (\bibinfo{year}{2023}).
\newblock


\bibitem[Kaddour et~al\mbox{.}(2023)]%
        {kaddour2023challenges}
\bibfield{author}{\bibinfo{person}{Jean Kaddour}, \bibinfo{person}{Joshua Harris}, \bibinfo{person}{Maximilian Mozes}, \bibinfo{person}{Herbie Bradley}, \bibinfo{person}{Roberta Raileanu}, {and} \bibinfo{person}{Robert McHardy}.} \bibinfo{year}{2023}\natexlab{}.
\newblock \showarticletitle{Challenges and applications of large language models}.
\newblock \bibinfo{journal}{\emph{arXiv preprint arXiv:2307.10169}} (\bibinfo{year}{2023}).
\newblock


\bibitem[Lannelongue et~al\mbox{.}(2021)]%
        {lannelongue2021green}
\bibfield{author}{\bibinfo{person}{Lo{\"\i}c Lannelongue}, \bibinfo{person}{Jason Grealey}, {and} \bibinfo{person}{Michael Inouye}.} \bibinfo{year}{2021}\natexlab{}.
\newblock \showarticletitle{Green algorithms: quantifying the carbon footprint of computation}.
\newblock \bibinfo{journal}{\emph{Advanced science}} (\bibinfo{year}{2021}).
\newblock


\bibitem[Lewis et~al\mbox{.}(2020)]%
        {lewis2019bart}
\bibfield{author}{\bibinfo{person}{Mike Lewis}, \bibinfo{person}{Yinhan Liu}, \bibinfo{person}{Naman Goyal}, \bibinfo{person}{Marjan Ghazvininejad}, \bibinfo{person}{Abdelrahman Mohamed}, \bibinfo{person}{Omer Levy}, \bibinfo{person}{Veselin Stoyanov}, {and} \bibinfo{person}{Luke Zettlemoyer}.} \bibinfo{year}{2020}\natexlab{}.
\newblock \showarticletitle{BART: Denoising Sequence-to-Sequence Pre-training for Natural Language Generation, Translation, and Comprehension}. In \bibinfo{booktitle}{\emph{Proceedings of the 58th ACL}}. \bibinfo{pages}{7871--7880}.
\newblock


\bibitem[Li et~al\mbox{.}(2022)]%
        {li2022survey}
\bibfield{author}{\bibinfo{person}{Qian Li}, \bibinfo{person}{Hao Peng}, \bibinfo{person}{Jianxin Li}, \bibinfo{person}{Congying Xia}, \bibinfo{person}{Renyu Yang}, \bibinfo{person}{Lichao Sun}, \bibinfo{person}{Philip~S Yu}, {and} \bibinfo{person}{Lifang He}.} \bibinfo{year}{2022}\natexlab{}.
\newblock \showarticletitle{A Survey on Text Classification: From Traditional to Deep Learning}.
\newblock \bibinfo{journal}{\emph{ACM Transactions on Intelligent Systems and Technology}} (\bibinfo{year}{2022}).
\newblock


\bibitem[Liang et~al\mbox{.}(2022)]%
        {holistic}
\bibfield{author}{\bibinfo{person}{Percy Liang}, \bibinfo{person}{Rishi Bommasani}, \bibinfo{person}{Tony Lee}, \bibinfo{person}{Dimitris Tsipras}, \bibinfo{person}{Dilara Soylu}, \bibinfo{person}{Michihiro Yasunaga}, \bibinfo{person}{Yian Zhang}, \bibinfo{person}{Deepak Narayanan}, \bibinfo{person}{Yuhuai Wu}, \bibinfo{person}{Ananya Kumar}, {et~al\mbox{.}}} \bibinfo{year}{2022}\natexlab{}.
\newblock \showarticletitle{Holistic evaluation of language models}.
\newblock \bibinfo{journal}{\emph{arXiv preprint arXiv:2211.09110}} (\bibinfo{year}{2022}).
\newblock


\bibitem[Liu et~al\mbox{.}(2019)]%
        {liu2019roberta}
\bibfield{author}{\bibinfo{person}{Yinhan Liu}, \bibinfo{person}{Myle Ott}, \bibinfo{person}{Naman Goyal}, \bibinfo{person}{Jingfei Du}, \bibinfo{person}{Mandar Joshi}, \bibinfo{person}{Danqi Chen}, \bibinfo{person}{Omer Levy}, \bibinfo{person}{Mike Lewis}, \bibinfo{person}{Luke Zettlemoyer}, {and} \bibinfo{person}{Veselin Stoyanov}.} \bibinfo{year}{2019}\natexlab{}.
\newblock \showarticletitle{Roberta: A robustly optimized bert pretraining approach}.
\newblock \bibinfo{journal}{\emph{arXiv preprint 1907.11692}} (\bibinfo{year}{2019}).
\newblock


\bibitem[Luhn(1957)]%
        {luhn1957statistical}
\bibfield{author}{\bibinfo{person}{Hans~Peter Luhn}.} \bibinfo{year}{1957}\natexlab{}.
\newblock \showarticletitle{A statistical approach to mechanized encoding and searching of literary information}.
\newblock \bibinfo{journal}{\emph{IBM Journal of R\&D}} (\bibinfo{year}{1957}).
\newblock


\bibitem[Luiz et~al\mbox{.}(2018)]%
        {cunha18}
\bibfield{author}{\bibinfo{person}{Washington Luiz}, \bibinfo{person}{Felipe Viegas}, \bibinfo{person}{Rafael Alencar}, \bibinfo{person}{Fernando Mour{\~a}o}, \bibinfo{person}{Thiago Salles}, \bibinfo{person}{D{\'a}rlinton Carvalho}, \bibinfo{person}{Marcos~Andre Gon{\c{c}}alves}, {and} \bibinfo{person}{Leonardo Rocha}.} \bibinfo{year}{2018}\natexlab{}.
\newblock \showarticletitle{A feature-oriented sentiment rating for mobile app reviews}. In \bibinfo{booktitle}{\emph{Proceedings of the 2018 World Wide Web Conference}}. \bibinfo{pages}{1909--1918}.
\newblock


\bibitem[Ma et~al\mbox{.}(2021)]%
        {adhocretrieval}
\bibfield{author}{\bibinfo{person}{Zhengyi Ma}, \bibinfo{person}{Zhicheng Dou}, \bibinfo{person}{Wei Xu}, \bibinfo{person}{Xinyu Zhang}, \bibinfo{person}{Hao Jiang}, \bibinfo{person}{Zhao Cao}, {and} \bibinfo{person}{Ji-Rong Wen}.} \bibinfo{year}{2021}\natexlab{}.
\newblock \showarticletitle{Pre-training for ad-hoc retrieval: hyperlink is also you need}. In \bibinfo{booktitle}{\emph{Proceedings of the 30th ACM CIKM}}. \bibinfo{pages}{1212--1221}.
\newblock


\bibitem[MacAvaney et~al\mbox{.}(2020)]%
        {documentRanking}
\bibfield{author}{\bibinfo{person}{Sean MacAvaney}, \bibinfo{person}{Franco~Maria Nardini}, \bibinfo{person}{Raffaele Perego}, \bibinfo{person}{Nicola Tonellotto}, \bibinfo{person}{Nazli Goharian}, {and} \bibinfo{person}{Ophir Frieder}.} \bibinfo{year}{2020}\natexlab{}.
\newblock \showarticletitle{Efficient Document Re-Ranking for Transformers by Precomputing Term Representations}. In \bibinfo{booktitle}{\emph{SIGIR'20}}.
\newblock
\showISBNx{9781450380164}


\bibitem[Matsubara et~al\mbox{.}(2020)]%
        {questionAnswering20}
\bibfield{author}{\bibinfo{person}{Yoshitomo Matsubara}, \bibinfo{person}{Thuy Vu}, {and} \bibinfo{person}{Alessandro Moschitti}.} \bibinfo{year}{2020}\natexlab{}.
\newblock \showarticletitle{Reranking for Efficient Transformer-Based Answer Selection}. In \bibinfo{booktitle}{\emph{ACM SIGIR}} (China). \bibinfo{numpages}{4}~pages.
\newblock
\showISBNx{9781450380164}


\bibitem[Muennighoff et~al\mbox{.}(2022)]%
        {muennighoff2022crosslingual}
\bibfield{author}{\bibinfo{person}{Niklas Muennighoff}, \bibinfo{person}{Thomas Wang}, \bibinfo{person}{Lintang Sutawika}, \bibinfo{person}{Adam Roberts}, \bibinfo{person}{Stella Biderman}, \bibinfo{person}{Teven~Le Scao}, \bibinfo{person}{M~Saiful Bari}, \bibinfo{person}{Sheng Shen}, \bibinfo{person}{Zheng-Xin Yong}, \bibinfo{person}{Hailey Schoelkopf}, {et~al\mbox{.}}} \bibinfo{year}{2022}\natexlab{}.
\newblock \showarticletitle{Crosslingual generalization through multitask finetuning}.
\newblock \bibinfo{journal}{\emph{arXiv preprint arXiv:2211.01786}} (\bibinfo{year}{2022}).
\newblock


\bibitem[Pasin et~al\mbox{.}(2025)]%
        {pasin2025quantumclef}
\bibfield{author}{\bibinfo{person}{Andrea Pasin}, \bibinfo{person}{Washington Cunha}, \bibinfo{person}{Maurizio~Ferrari Dacrema}, \bibinfo{person}{Paolo Cremonesi}, \bibinfo{person}{Marcos Goncalves}, {and} \bibinfo{person}{Nicola Ferro}.} \bibinfo{year}{2025}\natexlab{}.
\newblock \showarticletitle{QuantumCLEF-Quantum Computing at CLEF}. In \bibinfo{booktitle}{\emph{European Conference on Information Retrieval}}. Springer.
\newblock


\bibitem[Pasin et~al\mbox{.}(2024)]%
        {pasinictir24}
\bibfield{author}{\bibinfo{person}{Andrea Pasin}, \bibinfo{person}{Washington Cunha}, \bibinfo{person}{Marcos Goncalves}, {and} \bibinfo{person}{Nicola Ferro}.} \bibinfo{year}{2024}\natexlab{}.
\newblock \showarticletitle{A Quantum Annealing Instance Selection Approach for Efficient and Effective Transformer Fine-Tuning}. In \bibinfo{booktitle}{\emph{Proceedings of the 2024 ACM SIGIR ICTIR}}.
\newblock


\bibitem[Pasin et~al\mbox{.}(2022)]%
        {pasin2022quantum}
\bibfield{author}{\bibinfo{person}{Andrea Pasin}, \bibinfo{person}{Washington Cunha}, \bibinfo{person}{Marcos~Andr{\'e} Gon{\c{c}}alves}, \bibinfo{person}{Nicola Ferro}, {et~al\mbox{.}}} \bibinfo{year}{2022}\natexlab{}.
\newblock \showarticletitle{A Quantum Annealing-Based Instance Selection Approach for Transformer Fine-Tuning}. In \bibinfo{booktitle}{\emph{the 14th Italian Information Retrieval Workshop}}.
\newblock


\bibitem[Sebastiani(2002)]%
        {sebastiani02}
\bibfield{author}{\bibinfo{person}{Fabrizio Sebastiani}.} \bibinfo{year}{2002}\natexlab{}.
\newblock \showarticletitle{Machine learning in automated text categorization}.
\newblock \bibinfo{journal}{\emph{ACM Comput. Surv.}} \bibinfo{volume}{34}, \bibinfo{number}{1} (\bibinfo{year}{2002}), \bibinfo{pages}{1--47}.
\newblock


\bibitem[Siino et~al\mbox{.}(2024)]%
        {siino2024text}
\bibfield{author}{\bibinfo{person}{Marco Siino}, \bibinfo{person}{Ilenia Tinnirello}, {and} \bibinfo{person}{Marco La~Cascia}.} \bibinfo{year}{2024}\natexlab{}.
\newblock \showarticletitle{The Text Classification Pipeline: Starting Shallow going Deeper}.
\newblock \bibinfo{journal}{\emph{arXiv preprint arXiv:2501.00174}} (\bibinfo{year}{2024}).
\newblock


\bibitem[Sokolova and Lapalme(2009)]%
        {Sokolova}
\bibfield{author}{\bibinfo{person}{Marina Sokolova} {and} \bibinfo{person}{Guy Lapalme}.} \bibinfo{year}{2009}\natexlab{}.
\newblock \showarticletitle{A Systematic Analysis of Performance Measures for Classification Tasks}.
\newblock \bibinfo{journal}{\emph{Information Processing \& Management ({IP\&M})}} \bibinfo{volume}{45}, \bibinfo{number}{4} (\bibinfo{date}{July} \bibinfo{year}{2009}), \bibinfo{pages}{427--437}.
\newblock
\showISSN{0306-4573}
\urldef\tempurl%
\url{https://doi.org/10.1016/j.ipm.2009.03.002}
\showDOI{\tempurl}


\bibitem[Spirling(2023)]%
        {spirling2023open}
\bibfield{author}{\bibinfo{person}{Arthur Spirling}.} \bibinfo{year}{2023}\natexlab{}.
\newblock \showarticletitle{Why open-source generative AI models are an ethical way forward for science}.
\newblock \bibinfo{journal}{\emph{Nature}} \bibinfo{volume}{616}, \bibinfo{number}{7957} (\bibinfo{year}{2023}), \bibinfo{pages}{413--413}.
\newblock


\bibitem[Team et~al\mbox{.}(2024)]%
        {team2024gemma}
\bibfield{author}{\bibinfo{person}{Gemma Team}, \bibinfo{person}{Morgane Riviere}, \bibinfo{person}{Shreya Pathak}, \bibinfo{person}{Pier~Giuseppe Sessa}, \bibinfo{person}{Cassidy Hardin}, \bibinfo{person}{Surya Bhupatiraju}, \bibinfo{person}{L{\'e}onard Hussenot}, \bibinfo{person}{Thomas Mesnard}, \bibinfo{person}{Bobak Shahriari}, \bibinfo{person}{Alexandre Ram{\'e}}, {et~al\mbox{.}}} \bibinfo{year}{2024}\natexlab{}.
\newblock \showarticletitle{Gemma 2: Improving open language models at a practical size}.
\newblock \bibinfo{journal}{\emph{arXiv preprint arXiv:2408.00118}} (\bibinfo{year}{2024}).
\newblock


\bibitem[Touvron et~al\mbox{.}(2023)]%
        {touvron2023}
\bibfield{author}{\bibinfo{person}{Hugo Touvron}, \bibinfo{person}{Thibaut Lavril}, \bibinfo{person}{Gautier Izacard}, \bibinfo{person}{Xavier Martinet}, \bibinfo{person}{Marie-Anne Lachaux}, \bibinfo{person}{Timothée Lacroix}, \bibinfo{person}{Baptiste Rozière}, \bibinfo{person}{Naman Goyal}, \bibinfo{person}{Eric Hambro}, \bibinfo{person}{Faisal Azhar}, \bibinfo{person}{Aurelien Rodriguez}, \bibinfo{person}{Armand Joulin}, \bibinfo{person}{Edouard Grave}, {and} \bibinfo{person}{Guillaume Lample}.} \bibinfo{year}{2023}\natexlab{}.
\newblock \bibinfo{title}{LLaMA: Open and Efficient Foundation Language Models}.
\newblock
\newblock
\showeprint[arxiv]{2302.13971}~[cs.CL]
\urldef\tempurl%
\url{https://arxiv.org/abs/2302.13971}
\showURL{%
\tempurl}


\bibitem[Uppaal et~al\mbox{.}(2023)]%
        {uppaal2023fine}
\bibfield{author}{\bibinfo{person}{Rheeya Uppaal}, \bibinfo{person}{Junjie Hu}, {and} \bibinfo{person}{Yixuan Li}.} \bibinfo{year}{2023}\natexlab{}.
\newblock \showarticletitle{Is fine-tuning needed? pre-trained language models are near perfect for out-of-domain detection}.
\newblock \bibinfo{journal}{\emph{arXiv preprint arXiv:2305.13282}} (\bibinfo{year}{2023}).
\newblock


\bibitem[Wang et~al\mbox{.}(2020)]%
        {wang2020minilm}
\bibfield{author}{\bibinfo{person}{Wenhui Wang}, \bibinfo{person}{Furu Wei}, \bibinfo{person}{Li Dong}, \bibinfo{person}{Hangbo Bao}, \bibinfo{person}{Nan Yang}, {and} \bibinfo{person}{Ming Zhou}.} \bibinfo{year}{2020}\natexlab{}.
\newblock \showarticletitle{Minilm: Deep self-attention distillation for task-agnostic compression of pre-trained transformers}.
\newblock \bibinfo{journal}{\emph{Advances in NeurIPS}}  \bibinfo{volume}{33} (\bibinfo{year}{2020}), \bibinfo{pages}{5776--5788}.
\newblock


\bibitem[Xu et~al\mbox{.}(2024)]%
        {xu2024intelligent}
\bibfield{author}{\bibinfo{person}{Kangming Xu}, \bibinfo{person}{Huiming Zhou}, \bibinfo{person}{Haotian Zheng}, \bibinfo{person}{Mingwei Zhu}, {and} \bibinfo{person}{Qi Xin}.} \bibinfo{year}{2024}\natexlab{}.
\newblock \showarticletitle{Intelligent Classification and Personalized Recommendation of E-commerce Products Based on Machine Learning}.
\newblock \bibinfo{journal}{\emph{arXiv preprint arXiv:2403.19345}} (\bibinfo{year}{2024}).
\newblock


\bibitem[Yadan(2019)]%
        {Yadan2019Hydra}
\bibfield{author}{\bibinfo{person}{Omry Yadan}.} \bibinfo{year}{2019}\natexlab{}.
\newblock \bibinfo{title}{Hydra - A framework for elegantly configuring complex applications}.
\newblock \bibinfo{howpublished}{Github}.
\newblock
\urldef\tempurl%
\url{https://github.com/facebookresearch/hydra}
\showURL{%
\tempurl}


\bibitem[Yang et~al\mbox{.}(2019)]%
        {yang2019xlnet}
\bibfield{author}{\bibinfo{person}{Zhilin Yang}, \bibinfo{person}{Zihang Dai}, \bibinfo{person}{Yiming Yang}, \bibinfo{person}{Jaime Carbonell}, \bibinfo{person}{Russ~R Salakhutdinov}, {and} \bibinfo{person}{Quoc~V Le}.} \bibinfo{year}{2019}\natexlab{}.
\newblock \showarticletitle{XLNet: Generalized Autoregressive Pretraining for Language Understanding}. In \bibinfo{booktitle}{\emph{NIPS}}, Vol.~\bibinfo{volume}{32}. \bibinfo{pages}{5754--5764}.
\newblock


\bibitem[Zhang et~al\mbox{.}(2024)]%
        {zhang2024benchmarking}
\bibfield{author}{\bibinfo{person}{Tianyi Zhang}, \bibinfo{person}{Faisal Ladhak}, \bibinfo{person}{Esin Durmus}, \bibinfo{person}{Percy Liang}, \bibinfo{person}{Kathleen McKeown}, {and} \bibinfo{person}{Tatsunori~B Hashimoto}.} \bibinfo{year}{2024}\natexlab{}.
\newblock \showarticletitle{Benchmarking large language models for news summarization}.
\newblock \bibinfo{journal}{\emph{Transactions of the ACL (TACL))}}  \bibinfo{volume}{12} (\bibinfo{year}{2024}), \bibinfo{pages}{39--57}.
\newblock


\bibitem[Zhang and Wang(2023)]%
        {zhang2023prompt}
\bibfield{author}{\bibinfo{person}{Zizhuo Zhang} {and} \bibinfo{person}{Bang Wang}.} \bibinfo{year}{2023}\natexlab{}.
\newblock \showarticletitle{Prompt learning for news recommendation}. In \bibinfo{booktitle}{\emph{Proceedings of the 46th International ACM SIGIR}}. \bibinfo{pages}{227--237}.
\newblock


\bibitem[Zhao et~al\mbox{.}(2024)]%
        {zhao2024advancing}
\bibfield{author}{\bibinfo{person}{Hang Zhao}, \bibinfo{person}{Qile~P Chen}, \bibinfo{person}{Yijing~Barry Zhang}, {and} \bibinfo{person}{Gang Yang}.} \bibinfo{year}{2024}\natexlab{}.
\newblock \showarticletitle{Advancing Single-and Multi-task Text Classification through Large Language Model Fine-tuning}.
\newblock \bibinfo{journal}{\emph{arXiv preprint arXiv:2412.08587}} (\bibinfo{year}{2024}).
\newblock


\bibitem[Zhao et~al\mbox{.}(2023)]%
        {zhao2023survey}
\bibfield{author}{\bibinfo{person}{Wayne~Xin Zhao}, \bibinfo{person}{Kun Zhou}, \bibinfo{person}{Junyi Li}, \bibinfo{person}{Tianyi Tang}, \bibinfo{person}{Xiaolei Wang}, \bibinfo{person}{Yupeng Hou}, \bibinfo{person}{Yingqian Min}, \bibinfo{person}{Beichen Zhang}, \bibinfo{person}{Junjie Zhang}, \bibinfo{person}{Zican Dong}, {et~al\mbox{.}}} \bibinfo{year}{2023}\natexlab{}.
\newblock \showarticletitle{A survey of large language models}.
\newblock \bibinfo{journal}{\emph{arXiv preprint arXiv:2303.18223}} \bibinfo{volume}{1}, \bibinfo{number}{2} (\bibinfo{year}{2023}).
\newblock


\end{thebibliography}

\end{document}